\documentclass[runningheads]{llncs}
\usepackage[T1]{fontenc}
\usepackage{url}
\usepackage{hyperref}
\usepackage{graphicx}
\usepackage{natbib}  
\usepackage{geometry}

\usepackage{xcolor}  
\usepackage{hyperref}  
\makeatletter
\def\blfootnote{\gdef\@thefnmark{}\@footnotetext}
\makeatother

\hypersetup{
    colorlinks=true,  
    linkcolor=blue,  
    urlcolor=blue,  
    citecolor=blue,
    pdfborder={0 0 0},
}

\title{AI Thinking as a Meaning-Centered Framework: Reimagining Language Technologies Through Community Agency}
\titlerunning{AI Thinking, Language Technologies and Community Agency}
\author{Jose F Quesada\inst{1}\orcidID{0000-0001-7458-5855}}
\authorrunning{J. F. Quesada}
\institute{Dpt. of Computer Science and Artificial Intelligence, University of Seville, Spain 
\email{jquesada@us.es}\\
\url{https://www.cs.us.es/} \\
\url{https://www.linkedin.com/in/jose-f-quesada/}
}

\begin{document}

\maketitle

\blfootnote{\vspace{0.5cm}

	{\bf LT4All 2025}. Language Technologies for All - 2025.
	{\emph Advancing Humanism through Language Technologies.}
Paris (FR), UNESCO Headquarters, 24-26 February 2025}

\begin{abstract}
While language technologies have advanced significantly, current approaches fail to address the complex sociocultural dimensions of linguistic preservation. AI Thinking proposes a meaning-centered framework that would transform technological development from creating tools FOR communities to co-creating solutions WITH them. This approach recognizes that meaningful solutions emerge through the interplay of cultural understanding, community agency, and technological innovation. The proposal articulates a holistic methodology and a five-layer technological ecosystem where communities maintain control over their linguistic and cultural knowledge representation. This systematic integration of community needs, cultural preservation, and advanced capabilities could revolutionize how we approach linguistic diversity preservation in the digital age.

\end{abstract}


\section{Introduction}

\subsection{Context and Motivation}

Language technologies have experienced significant advances through foundation models, demonstrating unprecedented capabilities in language processing and generation \citep{Bommasani_etal_2021}. However, these technological developments have not benefited all linguistic communities equally. A substantial portion of the world's linguistic communities remain marginalized or entirely excluded from these innovations, revealing systemic limitations in current technological approaches.

\begin{quote}
``Language technologies contribute to promoting multilingualism and linguistic diversity
around the world. However, only a very small
number of the over 7000 languages of the
world are represented in the rapidly evolving
language technologies and applications. \citep[p.~6282]{Joshi_etal_2020}
\end{quote}

The dominance of foundation models in language technology introduces complex technical and cultural challenges. Contemporary approaches often transform languages into computational artifacts, critically separating them from their rich cultural contexts and community knowledge systems \citep{Munk_etal_2022}. The representation of meaning through vector embeddings, while computationally efficient, problematically reduces semantic complexity to mere statistical correlations. Moreover, the standardization of language technology development through foundation models generates potential vulnerabilities for cultural and linguistic preservation \citep{Noble_2018}.

These challenges become particularly acute as digital technologies increasingly mediate global communication. Recent indigenous language technology research \citep{Mager_etal_2021} reveals both promising opportunities and significant limitations in developing culturally appropriate technological solutions. The persistent technological gap between supported and unsupported languages \citep{Melero_2018} underscores the urgent need for innovative approach to language technology development.

The integration of traditional knowledge systems with modern language technologies demands nuanced consideration of technical and cultural dimensions: 

\begin{quote}
``
Amidst globalization and the Fourth IndustrialRevolution, indigenous language preservation and promo-tion gain paramount significance. These languages encap-sulate rich cultural, historical, and ecological value,embodying unique knowledge systems and identities.Nevertheless, challenges stemming from historical, social,resource, and institutional constraints impede preservation,with digital media technology thus emerging as a pivotal toolin this endeavour.   \citep[p.~35]{Ajani_etal_2024}
\end{quote}

While foundation models offer powerful computational capabilities, their application must fundamentally account for preserving cultural practices and linguistic diversity. This context necessitates frameworks that can effectively synthesize computational advancement with cultural preservation, maintaining the sophistication required for contemporary digital communication.

Emerging research suggests that addressing these challenges requires approaches extending beyond purely technical solutions \citep{Crawford_2021}. The development of language technologies must simultaneously consider computational efficiency and cultural preservation, especially in contexts where traditional knowledge systems are intrinsic to community identity and communication \citep{Reyhner_Lockard_2009}.

\subsection{State of the Art and Related Work}

Current approaches to language preservation through technology can be categorized into several primary streams.

Foundation models and neural architectures represent the dominant paradigm, offering sophisticated language processing capabilities \citep{Bommasani_etal_2021}. These approaches have demonstrated effectiveness for well-resourced languages but encounter significant limitations when applied to low-resource linguistic contexts \citep{Ranathunga_etal_2023}.

Community-driven approaches have emerged as a crucial alternative paradigm. \citep{Bird_2022} documents efforts to develop {\it third spaces} where technological innovation and community knowledge productively intersect. These initiatives emphasize local control over language resources and development processes, though they frequently confront resource and scalability constraints.

Participatory design methodologies in language technology development have shown promising potential for addressing community needs. \citep{Hutchinson_etal_2021} articulates frameworks for accountability in dataset creation and technological development, emphasizing sustained community involvement. However, these approaches often struggle to balance meaningful community participation with complex technical requirements.

\begin{quote}
``According to Clifford Geertz, the purpose of anthropology is not to explain culture but to explicate it. That should cause us to rethink our relationship with machine learning. It is, we contend, perfectly possible that machine learning algorithms, which are unable to explain, and could even be unexplainable themselves, can still be of critical use in a process of explication.'' \citep[]{Hutchinson_etal_2021}
\end{quote}

Several fundamental limitations characterize current technological approaches. First, existing models typically require extensive digital language data, which many communities lack. Second, standard computational methods frequently fail to capture the nuanced contextual and cultural dimensions of language use \citep{Munk_etal_2022}. Third, the inherent technical complexity of current solutions can create substantial barriers to community participation and control.

Research has systematically identified critical gaps in existing approaches. \citep{Joshi_etal_2020} documented persistent systematic biases in language technology development, particularly affecting low-resource languages. The integration of traditional knowledge systems with modern technological frameworks remains challenging, with few existing methodologies successfully bridging this epistemological divide \citep{Bijker_etal_1987}. 

Consequently, sustainable models for community-driven technological development remain nascent.
Recent scholarship increasingly suggests the necessity of approaches that can simultaneously leverage modern language technologies' capabilities while centering community-driven development practices. This requires frameworks capable of addressing both technical challenges and cultural preservation needs, ensuring genuine community agency in technological development \citep{Latour_2005}. 

\subsection{Language Technologies and the Complexity of Human Communication}

Language represents a complex, multidimensional phenomenon deeply rooted in biological, psychological, social, and cultural domains. Contemporary technological approaches, despite their computational power, often reduce this rich, nuanced reality to simplified computational representations \citep{Munk_etal_2022}. This reductive approach presents significant challenges for preserving the intricate richness of human linguistic expression and cultural transmission systems.

Anthropological linguistics and cognitive science research emphasizes language's profound role in shaping human experience across multiple dimensions \citep{Suchman_2006}. Language simultaneously functions as a biological capacity, cognitive system, social practice, and cultural repository. Existing technological approaches predominantly focus on surface-level patterns and statistical regularities, potentially overlooking deeper linguistic phenomena.

The standardization of language processing through foundation models, while technically efficient, risks oversimplifying this inherent complexity. Current approaches may inadvertently promote linguistic homogenization by prioritizing computational efficiency over cultural and social dimensions. This technological standardization impacts not only language processing but fundamentally shapes how communities interact with and transmit their linguistic heritage.

The challenges extend beyond mere technical considerations. \citep{Crawford_2021} reveals how current approaches can disconnect languages from their ecological contexts—the intricate web of social practices, cultural knowledge, and community relationships that imbue language with meaning and vitality. This disconnection particularly impacts communities whose linguistic practices are deeply intertwined with traditional knowledge systems and cultural practices \citep{Ajani_etal_2024}.

Recent studies highlight specific risks in technological simplification. \citep{Noble_2018} demonstrates how algorithmic approaches can inadvertently reinforce existing power structures and cultural biases. The emphasis on data-driven methodologies poses particular challenges in understanding and preserving subtle cultural and contextual language use dimensions \citep{Hutchinson_etal_2021}. Moreover, the abstraction of language into computational models risks losing what \citep{Latour_2005} describes as the rich network of associations constituting genuine linguistic practice.

These challenges necessitate approaches that can more comprehensively accommodate human language's full complexity. Such approaches must consider how technological solutions can support—rather than supplant—the rich, multifaceted biological, psychological, social, and cultural factors constituting genuine linguistic practice. This requires frameworks bridging computational efficiency and language's multidimensional preservation.

\subsection{AI Thinking: A Framework for Language Technology Development}

AI Thinking offers a methodological framework for addressing language's multidimensional nature through systematic integration of technological capabilities with cultural-semiotic processes \citep{Quesada_2024}. Inspired from decades of AI research, this framework establishes structured approaches for developing technologies that preserve and enhance the complexity of human linguistic practices.

The framework's core principles position meaning as the central axis guiding technological development, providing mechanisms for preserving the intricate interplay between cognitive, social, and cultural dimensions of language. This approach enables meaningful integration of community knowledge systems with technological development.
For language technologies, AI Thinking proposes specific methodological approaches. The framework's emphasis on cultural-semiotic processes provides mechanisms for integrating community knowledge systems with technological development. By recognizing agency as a fundamental dimension, it supports technologies that enhance rather than replace community practices.

The framework's hierarchical structure—incorporating core principles, structural dimensions, and operational skills—provides systematic methodologies for meaning preservation across computational processes. It ensures integration of community knowledge systems and maintenance of cultural authenticity in language processing \citep{Noble_2018}. Its meaning-centered design enables preservation of cultural context while leveraging technological capabilities \citep{Latour_2005}.
Through methodological rigor, AI Thinking supports developing solutions that combine computational approaches with traditional linguistic practices \citep{Munk_etal_2022}, offering systematic approaches for language technologies that serve community needs while preserving linguistic complexity.

\subsection{Paper Structure and Research Contributions}

This paper presents a meaning-centered approach to language technology development based on the AI Thinking framework. The proposal develops three primary contributions to the field of language technologies and cultural preservation.

Section 2 critically analyzes current challenges in language technology development, examining how foundation models and standardized approaches impact linguistic diversity and cultural preservation. Section 3 introduces the AI Thinking framework, establishing its theoretical foundations and hierarchical structure for meaning-centered technology development. Section 4 develops a comprehensive technological ecosystem grounded in AI Thinking principles. This five-layer architecture integrates knowledge representation, intelligence, interface, integration, and preservation capabilities. The section establishes methodological guidelines for implementing meaning-centered language technologies while maintaining community agency.
Section 5 examines implementation challenges and potential research directions. The analysis addresses technical requirements for preserving linguistic complexity, methodological approaches for community participation, and governance frameworks for sustainable development. 

The proposal address in this paper acknowledges the inherent complexity of bridging technological innovation with cultural preservation. While proposing a novel framework, we recognize this as an exploratory intervention rather than a definitive solution. Our aim is to contribute to a more nuanced, culturally responsive approach to language technology development.

\section{Motivation and Current Challenges}

Recent advances in language technologies, particularly through foundation models, have demonstrated unprecedented capabilities in processing and generating human language \citep{Bommasani_etal_2021}. However, this technological progress masks a fundamental tension between computational advancement and linguistic diversity preservation. A significant portion of the world's linguistic communities remains underserved or entirely excluded from these developments \citep{Bird_2020, Bird_2022}, reflecting deeper systemic challenges in our approach to language technology development.

The emergence of foundation models represents a critical juncture in the evolution of language technologies. As Bommasani et al. note:

\begin{quote}
``Foundation models have led to an unprecedented level of homogenization: Almost all state-of-the-art NLP models are now adapted from one of a few foundation models [...] While this homogenization produces extremely high leverage (any improvements in the foundation models can lead to immediate benefits across all of NLP), it is also a liability [...] We are also beginning to see a homogenization across research communities.'' \citep[p.~5]{Bommasani_etal_2021}.
\end{quote}

This homogenization presents a paradox: while it enables rapid technological advancement through standardized approaches, it simultaneously threatens linguistic and cultural diversity by imposing uniform computational paradigms across fundamentally different linguistic and cultural contexts. The current paradigm in language technology development, predominantly shaped by this homogenizing computational perspective, prioritizes processing efficiency and algorithmic optimization over cultural preservation.

Three critical risks emerge from this trajectory:

\begin{itemize}

\item First, the transformation of languages into computational artifacts risks divorcing them from their cultural contexts and community knowledge systems. This separation undermines the essence of language as a medium for cultural transmission and meaning-making \citep{Munk_etal_2022}. 

\item Second, the prevalent approach of representing meaning through vector embeddings in high-dimensional spaces, while mathematically elegant and computationally efficient, reduces the rich complexity of semantic understanding to purely statistical correlations. 

\item Third, and perhaps most critically, the homogenization of language technology development through foundation models creates a single point of failure for cultural and linguistic preservation, potentially embedding and amplifying existing biases and power structures across all applications \citep{Bender_etal_2021}.
\end{itemize}

The urgency of addressing these challenges is amplified by the rapid digitalization of global communication. While technological advancement through foundation models is inevitable and potentially beneficial, its current trajectory threatens to marginalize traditional knowledge systems and cultural practices that have sustained linguistic communities for generations. This tension between technological progress and cultural preservation necessitates a fundamental reimagining of how we approach language technology development.

In response to these challenges, we propose a paradigm shift from a technology-centered to a meaning-centered approach, emphasizing three fundamental transitions:

\begin{enumerate}
    \item From homogenized processing to diverse meaning preservation, embracing the full richness of cultural-linguistic expression
    \item From standardized tool development to community empowerment, positioning speech communities as active creators rather than passive recipients
    \item From computational optimization to heritage preservation, ensuring technological advancement serves rather than supplants cultural knowledge systems
\end{enumerate}

This reconceptualization aligns with UNESCO's vision for advancing humanism through technology \citep{Sapignoli_2021}, while addressing the practical challenges of developing sustainable, culturally appropriate language technologies. The framework we propose seeks to bridge the gap between cutting-edge technological capabilities and traditional knowledge systems, creating solutions that are both technically sophisticated and culturally resonant.

The essence of this challenge lies in recognizing that meaningful technological solutions emerge not from homogenized computational optimization alone, but through the complex interplay of cultural understanding, community participation, and technological innovation \citep{Suchman_2006}. This perspective necessitates moving beyond the creation of standardized tools \emph{for} communities, towards the co-creation of diverse solutions \emph{with} communities, ensuring that technological advancement serves the genuine needs of linguistic preservation and cultural empowerment.

In synthesizing these challenges and risks, we identify two fundamental transitions that must guide future development:

\begin{itemize}
\item {\bf FROM technologies TO communities}; and 
\item {\bf FROM processing TO language understanding}.
\end{itemize}

These transitions embody three essential principles: 

\begin{itemize}
\item Meaning over processing (embracing cultural richness), 
\item Agency over tools (empowering communities as creators), and 
\item Heritage over efficiency (preserving cultural knowledge). 
\end{itemize}

The risks of failing to make these transitions are profound: the reduction of languages to mere datasets, the conflation of meaning with vector representations in latent spaces, and the homogenization of linguistic diversity through standardized computational approaches. While technical innovations like embeddings and foundation models represent important advances, their reductionist application threatens to diminish the rich tapestry of human linguistic and cultural expression. Our framework, therefore, seeks not to reject these technological advances, but to recontextualize them within a broader understanding of language as a living, cultural phenomenon that transcends purely computational representation.

\section{AI Thinking: A Conceptual Framework}

AI Thinking represents a systematic approach to leveraging the methodological insights gained from decades of artificial intelligence research to enhance human cognitive processes and improve technology development. This framework emerges from the convergence of three fundamental domains: education as a mechanism for knowledge transmission and cognitive development, intelligence as a complex set of cognitive capabilities, and artificial intelligence as a scientific-engineering endeavor to model and implement these capabilities \citep{Mitchell_2020}.

The framework's theoretical foundation rests on the recognition that meaningful technological advancement requires more than computational sophistication—it demands a deep understanding of how knowledge emerges, evolves, and transmits within cultural contexts. This perspective aligns with recent developments in cognitive science and artificial intelligence that emphasize the importance of meaning-making processes over pure information processing.

\begin{quote}
``The main objective of the AI Thinking approach is not to introduce specific AI content into the curriculum (this is something that is already being done within the approaches that we have previously presented as AI as a content or AI literacy). The challenges and opportunities of this approach derive from the potential to take advantage in the world of education of all the methodologies, good practices,
in short, all the results obtained during decades of work in AI focused on modelling cognitive capacities in computer systems. It is not a question of trying to model human cognitive capacities computationally. Quite the contrary. It is about being able to optimise, to improve some educational processes by considering the experiences in AI. And always assuming that human cognitive capacities
are very different from the capacities offered by current computational models used in AI.''\citep[p.~4315]{Quesada_2024}
\end{quote}

\subsection{Core Principles}

At its heart, AI Thinking posits that meaning emerges through dynamic cultural-semiotic processes rather than through computational processing alone. This fundamental principle recognizes that knowledge and understanding are inherently embedded in cultural contexts and social practices \citep{Suchman_2006}. The framework emphasizes how communities construct, validate, and transmit knowledge through complex interactions that transcend simple information exchange \citep{Riedl_2019, Marcus_Davis_2020, Miller_2022}.

The relationship between culture and cognition forms a central axis in AI Thinking. Culture shapes not only what we know but how we know—influencing our cognitive processes, our ways of reasoning, and our approaches to problem-solving. This cultural-cognitive interface becomes particularly crucial when considering how technological systems can support rather than supplant traditional knowledge systems.

{\bf AI Thinking is a conceptual framework that
systematically applies the methodological
lessons learned from decades of AI research
to enhance human cognitive processes and
improve technology development.}

\subsection{Hierarchical Structure}

The AI Thinking framework establishes a tripartite hierarchical architecture that integrates meaning-making processes with cognitive development and technological innovation. This structure emerges from the recognition that effective cognitive enhancement requires a sophisticated interplay between fundamental principles, structural elements, and operational capabilities. The framework's organization reflects the complex nature of human cognition while providing a systematic approach to its development and enhancement.

\begin{figure}[htbp]
    \centering
    \includegraphics[width=0.8\textwidth]{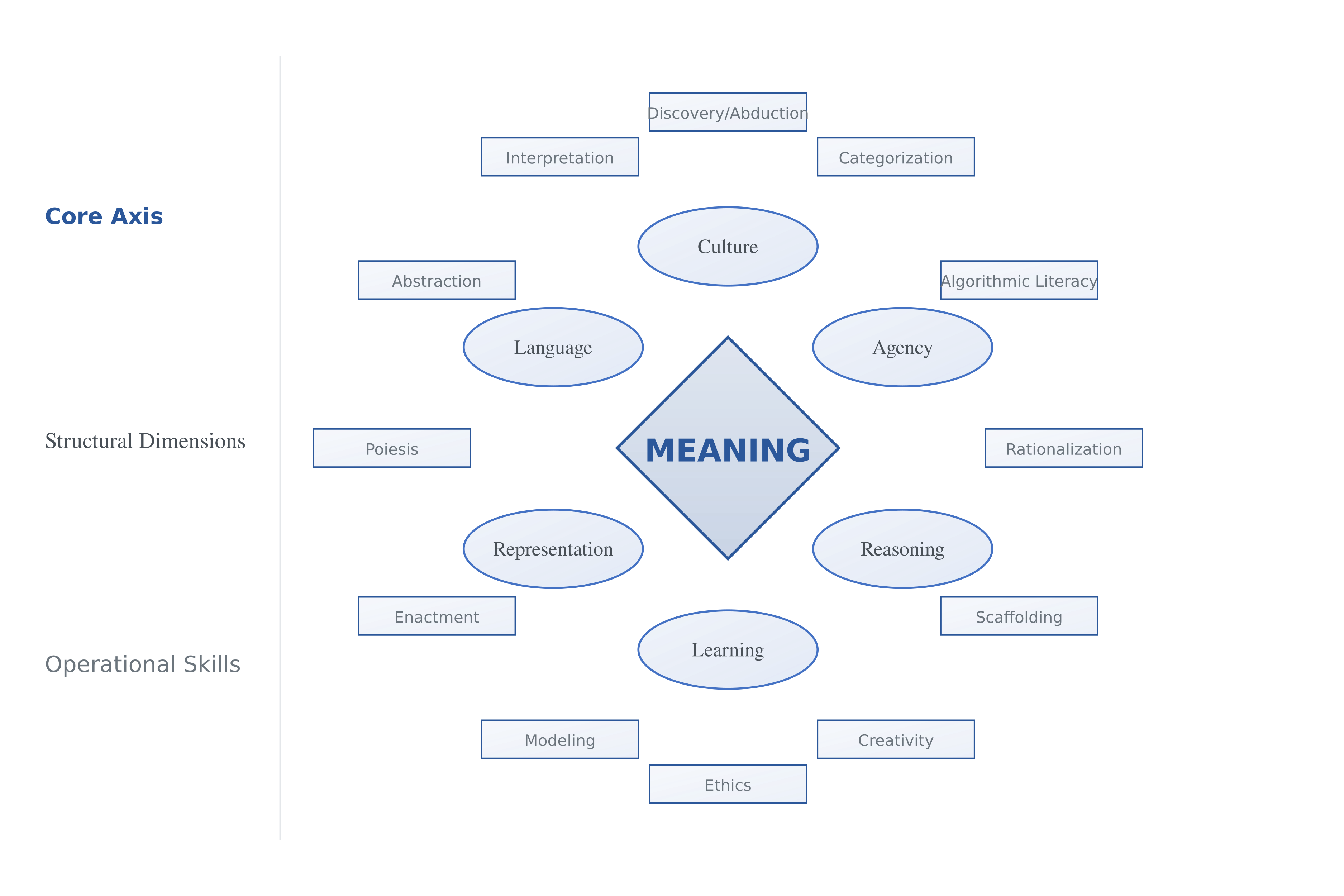}
    \caption{AI Thinking: Hierarchical framework}
    \label{fig:hierarchical-framework}
\end{figure}

\subsubsection{The Core Axis: Meaning as Foundational Principle}

At the heart of AI Thinking lies the concept of meaning, serving as the fundamental axis around which all other elements revolve. This positioning of meaning as the core principle represents a significant departure from traditional computational approaches that prioritize processing efficiency or algorithmic optimization. In the AI Thinking framework, meaning emerges through dynamic semiotic processes that integrate individual cognition with cultural contexts and collective knowledge systems. This radical emphasis on meaning acknowledges that genuine understanding and cognitive development cannot be reduced to mere information processing but must engage with the rich tapestry of human sense-making and cultural interpretation.

The centrality of meaning in this framework reflects contemporary insights from cognitive science, semiotics, and cultural anthropology, suggesting that human intelligence fundamentally operates through meaning-making processes rather than through purely computational mechanisms. This understanding has profound implications for how we approach cognitive development and technological enhancement of human capabilities.

\subsubsection{Structural Dimensions: The Architecture of Understanding}

The framework's second level comprises six structural dimensions that articulate the primary domains through which meaning manifests and operates. These dimensions do not function as isolated components but rather as interconnected aspects of a unified system of meaning-making and cognitive development.

\begin{itemize}
\item {\bf Culture}, as a structural dimension, encompasses the shared systems of meaning and practice that provide the context for all cognitive activity. This dimension recognizes that human understanding is inherently situated within cultural frameworks that shape both what we know and how we come to know it. The cultural dimension interacts dynamically with linguistic systems, which serve as the primary medium for the construction and transmission of meaning.

\item {\bf Language}, operating both as a structural dimension and a meaning-making tool, facilitates the complex processes of knowledge construction and social interaction. Through language, individuals access cultural knowledge, engage in collective meaning-making, and develop sophisticated cognitive capabilities. This dimension encompasses not only natural language but also formal symbolic systems and various modes of representation.

\item {\bf Agency} emerges as a crucial dimension that addresses the capacity for intentional action and self-directed learning within cultural-semiotic systems. This dimension recognizes that effective cognitive development requires active engagement and purposeful participation in meaning-making processes. Agency operates at both individual and collective levels, shaping how learners interact with knowledge systems and cultural contexts.

\item The {\bf Representation} dimension concerns the various ways in which knowledge and meaning are encoded, stored, and communicated. This encompasses both internal cognitive representations and external symbolic systems, recognizing that different forms of representation enable different types of understanding and cognitive processing.

\item {\bf Learning}, as a structural dimension, focuses on the processes through which new understanding is constructed and integrated into existing knowledge frameworks. This dimension emphasizes the dynamic nature of knowledge acquisition and the importance of meaningful connection-making in cognitive development.

\item The {\bf Reasoning} dimension addresses the cognitive mechanisms that enable systematic meaning-making and problem-solving. This encompasses both formal logical processes and more intuitive forms of understanding, recognizing that human reasoning operates through multiple, complementary modes of thought.
\end{itemize}

\subsubsection{Operational Skills: Implementing Cognitive Enhancement}

The framework's third level comprises twelve operational skills that represent the practical implementation of AI Thinking principles. These skills emerge from the interaction between the core meaning axis and the structural dimensions, providing concrete capabilities for cognitive enhancement and technological innovation.

\begin{itemize}
\item Rationalization and Discovery represent complementary approaches to knowledge development. {\bf Rationalization} enables systematic analysis and organization of knowledge, while {\bf Discovery}, incorporating {\bf Abductive reasoning}, facilitates the generation of novel hypotheses and explanatory frameworks. These foundational skills support both analytical and creative aspects of cognitive development.

\item The framework includes sophisticated cognitive operations through Categorization and Abstraction. {\bf Categorization} enables the organization of knowledge into meaningful conceptual structures, while {\bf Abstraction} facilitates the identification of essential patterns and principles across diverse domains. These skills support higher-order thinking and complex problem-solving capabilities.

\item Creative and procedural aspects of cognition are addressed through Creativity and Algorithmic Literacy. {\bf Creativity} encompasses the generation of novel solutions and perspectives, while {\bf Algorithmic Literacy} enables understanding of systematic problem-solving approaches. This pairing acknowledges the importance of both divergent and convergent thinking in cognitive development.

\item The framework incorporates advanced cognitive skills through Hierarchical Modeling and Poiesis. {\bf Hierarchical Modeling} enables the construction of sophisticated representational frameworks, while {\bf Poiesis} addresses the active creation of meaning through various modalities. These skills support complex knowledge construction and meaningful learning experiences.

\item Interpretation and Scaffolding skills focus on meaning-making and learning support. {\bf Interpretation} enables sophisticated analysis of signs and symbols, while {\bf Scaffolding} provides structured support for learning and development. These skills facilitate both independent learning and guided development.

\item The framework culminates in Enactment and Reflection skills, which connect understanding with practical application. {\bf Enactment} enables the implementation of knowledge in real-world contexts, while {\bf Ethics} supports critical examination of processes and outcomes. These skills ensure that cognitive development remains grounded in practical application while maintaining theoretical sophistication.
\end{itemize}

\subsubsection{Integration and Systemic Interaction}

The three levels of the AI Thinking framework—core axis, structural dimensions, and operational skills—function as an integrated system rather than as isolated components. This integration operates through multiple mechanisms of interaction, including vertical integration across levels, horizontal interaction within levels, and dynamic adaptation to varying contexts. The framework's systematic yet flexible nature enables its application across diverse educational and technological contexts while maintaining its fundamental focus on meaning as the central organizing principle.

\section{A Meaning-Centered Technological Framework for Language Preservation}

This section proposes a comprehensive framework that aims to bridge conceptual foundations with technological implementation, designed to ensure that meaning-making processes and community agency remain central to all aspects of development. Our proposal represents a systematic approach to translate the theoretical principles of AI Thinking into practical implementation while addressing the concrete challenges of language technology development.

The proposed framework suggests a departure from traditional approaches to language technology development, which often prioritize computational efficiency over cultural preservation. Instead, we envision an integrated ecosystem that would systematically implement the principles of AI Thinking while addressing the practical challenges of creating sustainable, community-driven language technologies. This approach recognizes that effective technological solutions should emerge from the synthesis of cultural understanding, community wisdom, and technical innovation.

\subsection{Conceptual Integration: From AI Thinking to Technological Implementation}

The integration of AI Thinking principles into technological development requires a fundamental shift in how we conceptualize and implement language technologies. We propose a transformation that would move beyond the traditional paradigm of creating tools that merely process linguistic data, towards developing systems that actively preserve and enhance meaning-making processes within cultural contexts. This transformation would operate across three fundamental dimensions:

First, we propose reconceptualizing language technologies as meaning preservation systems rather than mere processing tools. This perspective acknowledges that language is fundamentally a vehicle for cultural transmission and meaning-making \citep{Suchman_2006}, necessitating technological solutions that would preserve not just linguistic forms but the complex web of cultural associations and meanings they carry. This approach would directly address the risk of meaning reduction identified in current language technology developments \citep{Bender_etal_2021}.

Second, our framework suggests establishing a systematic connection between the structural dimensions of AI Thinking and concrete technological components. The cultural, linguistic, and agency dimensions of AI Thinking would find direct expression in technological systems through specialized layers designed to preserve and enhance these aspects. This alignment would ensure that technological development remains grounded in the fundamental principles of meaning-centered design while providing practical solutions for language preservation.

Third, the framework proposes implementing the operational skills identified in AI Thinking through specific technological mechanisms. Skills such as rationalization, discovery, and poiesis would be embedded within the technological architecture through specialized components that support these cognitive processes. This integration aims to ensure that the resulting technologies not only preserve linguistic content but actively support the cognitive processes through which communities construct and transmit meaning.

We propose implementing what we term {\it meaning-preservation protocols} – systematic approaches to ensuring that cultural significance and community agency are maintained throughout the technological development process. These protocols would address three critical aspects of integration:

\subsubsection{Cultural-Technical Synthesis}

The framework would establish mechanisms for translating cultural knowledge systems into technological implementations without reducing their complexity or richness. This synthesis acknowledges that effective language technologies must emerge from deep understanding of how communities construct and transmit meaning \citep{Bird_2020}. The technical architecture would incorporate specific components for capturing and preserving cultural context, ensuring that linguistic data remains connected to its broader cultural significance.

\subsubsection{Community-Driven Development}

The proposed integration framework would position communities as active creators rather than passive recipients of technology. This approach would implement the agency dimension of AI Thinking through specific technological mechanisms that enable communities to shape and control their language technologies. The resulting systems would be designed to evolve with community needs and practices, ensuring long-term sustainability and cultural relevance.

\subsubsection{Knowledge Integration Systems}

The framework envisions sophisticated mechanisms for integrating different forms of knowledge – traditional and computational, explicit and tacit, individual and collective. This integration reflects the AI Thinking principle that meaningful solutions emerge from the synthesis of diverse knowledge systems \citep{Marcus_Davis_2020}. The technical architecture would include specialized components for knowledge representation that preserve the richness of traditional understanding while leveraging modern computational capabilities.

\subsection{Methodological Guidelines for Community-Driven Development}

We propose a set of methodological guidelines aimed at systematically integrating community participation, cultural preservation, and technological innovation. These guidelines would establish structured approaches for developing language technologies that emerge from and serve community needs while maintaining scientific rigor and technical sophistication.

Our methodology would operate through what we term {\it cultural-technical synthesis cycles} – iterative processes designed to combine community knowledge with technological development in a systematic manner. This approach would address the fundamental challenge identified by \citep{Bird_2022} regarding the gap between computational advancement and community needs, while implementing the meaning-preservation principles established in our conceptual framework.

Community engagement would form the foundation of the methodology, but this engagement aims to go beyond traditional participatory design approaches. Instead, we propose implementing what \citep{Sapignoli_2021} describes as "deep participation" – a process where communities would actively shape not just the features of technological tools but the fundamental paradigms through which these tools operate. This approach would manifest through three interconnected methodological components:

Knowledge Integration Frameworks would serve as the primary mechanism for synthesizing community wisdom with technological capabilities. These frameworks would implement systematic approaches for documenting and preserving cultural knowledge while adapting technological solutions to community needs. The process would begin with comprehensive cultural-linguistic documentation, conducted through methods that respect and preserve the complex ways in which communities construct and transmit meaning. This documentation would extend beyond traditional linguistic annotation to capture what \citep{Suchman_2006} terms the "situated nature of meaning" – the complex web of cultural associations and practices through which language acquires significance.

The development process would follow what we term "meaning-centered iterative cycles," designed to extend traditional agile methodologies to incorporate cultural validation and meaning preservation at each stage. These cycles would implement specific protocols for ensuring that technological development aligns with community needs and cultural practices:

\begin{enumerate}
    \item Cultural Context Analysis: Systematic documentation of meaning-making processes within the community, including both explicit linguistic practices and implicit cultural knowledge.
    
    \item Technical Component Design: Development of technological solutions explicitly designed to preserve and enhance documented meaning-making processes.
    
    \item Community Validation: Structured processes for community evaluation and refinement of technological components, focusing on meaning preservation and cultural authenticity.
    
    \item Integration and Adaptation: Systematic incorporation of community feedback and cultural insights into technological implementations.
\end{enumerate}

The proposed evaluation frameworks would extend beyond traditional metrics of technical performance to incorporate what \citep{Crawford_2021} terms "cultural metrics" – measures designed to assess how effectively technological solutions preserve and enhance community meaning-making processes. These evaluation frameworks would operate through multiple dimensions:

Cultural Authenticity would assess how effectively technological solutions preserve and transmit cultural meaning. This dimension would examine not just linguistic accuracy but the broader cultural resonance of technological implementations. We propose evaluation protocols that would include community-defined metrics for assessing cultural preservation and meaning transmission.

Community Empowerment would measure the degree to which technological solutions enhance community agency in preserving and developing their linguistic heritage. This dimension would implement specific metrics for assessing community control over technological tools and their evolution.

Sustainability Assessment would examine the long-term viability of technological solutions within community contexts. This dimension would consider both technical sustainability and cultural sustainability, assessing how effectively solutions could be maintained and evolved by communities themselves. The assessment would follow frameworks developed by \citep{Munk_etal_2022} for evaluating the long-term impact of technological interventions in cultural contexts.

The proposed methodology incorporates specific protocols for knowledge transfer and capacity building, aimed at ensuring that communities develop the capabilities needed to maintain and evolve their technological solutions. These protocols would implement what \citep{Latour_2005} terms "knowledge networks" – systems through which technical expertise and cultural wisdom can be effectively synthesized and transmitted. This approach aims to ensure that technological development serves as a means of community empowerment rather than dependency.

Throughout all stages, the methodology would maintain a strong focus on what we term "meaning verification" – systematic processes designed to ensure that technological implementations accurately preserve and transmit cultural meaning. These verification processes would combine technical validation with cultural validation, implementing specific protocols for assessing how effectively technological solutions maintain the integrity of community meaning-making processes.

\subsection{A Five-Layer Technological Ecosystem}

Based on our conceptual framework and methodological guidelines, we propose a comprehensive technological ecosystem consisting of five interconnected layers, as illustrated in Figure~\ref{fig:tech-ecosystem}. Each layer is designed to address specific aspects of meaning preservation and community empowerment while maintaining technological sophistication. The proposed ecosystem would implement a vertical integration strategy, where each layer builds upon and supports the others, creating a cohesive system for language preservation and development.

\subsubsection{Knowledge Representation Layer}

\begin{figure}[htbp]
    \centering
    \includegraphics[width=0.85\textwidth]{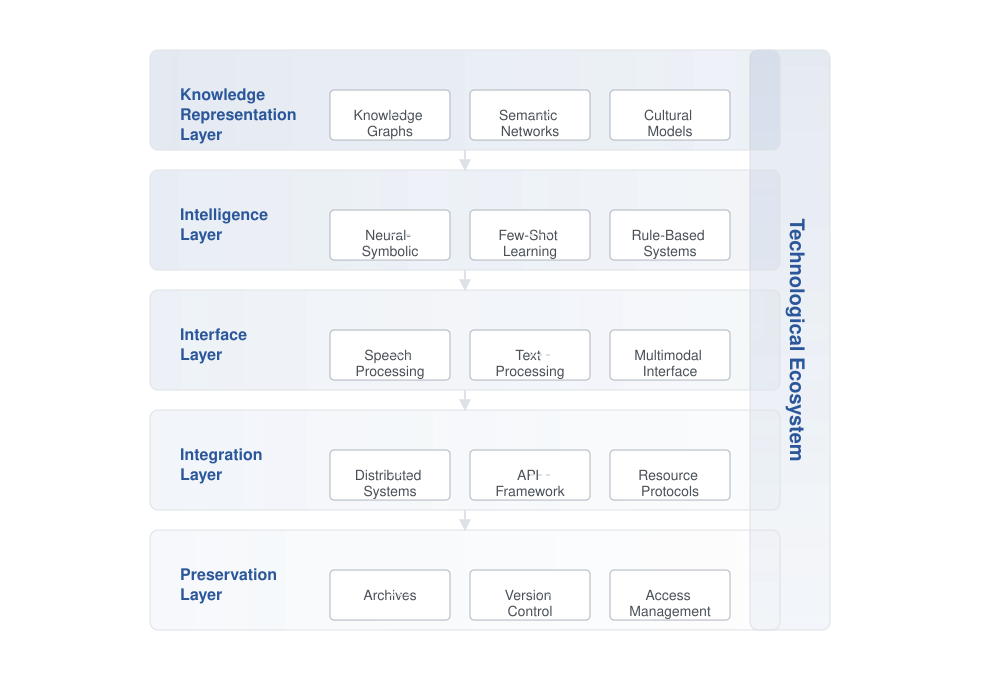}
    \caption{Five-Layer Technological Ecosystem for Language Preservation}
    \label{fig:tech-ecosystem}
\end{figure}

The foundational layer of our proposed ecosystem would focus on the sophisticated representation of linguistic and cultural knowledge. As shown in Figure~\ref{fig:tech-ecosystem}, this layer would comprise three primary components: Knowledge Graphs, Semantic Networks, and Cultural Models.

Knowledge Graphs would serve as the foundation for representing complex linguistic structures and relationships. We propose implementing these through specialized graph database technologies enhanced with linguistic annotation capabilities. These graphs would be designed to capture not only morphological patterns and syntactic structures but also the deep cultural contexts in which they operate \citep{Bird_2020}.

Semantic Networks would extend beyond traditional linguistic relationships to encompass cultural meaning systems. These networks would implement advanced RDF/OWL frameworks, designed to capture ontological relationships between linguistic concepts while preserving their cultural significance. The proposed approach would allow for the representation of meaning relationships that reflect community understanding rather than purely computational correlations.

Cultural Models would complete this layer by providing specialized frameworks for representing context-specific knowledge. These models would be designed to capture and preserve cultural nuances that might be lost in traditional linguistic representations, implementing what \citep{Suchman_2006} describes as "situated meaning structures."

\subsubsection{Intelligence Layer}

Building upon the knowledge representation foundation, the Intelligence Layer would implement hybrid approaches combining advanced AI capabilities with community wisdom. As depicted in Figure~\ref{fig:tech-ecosystem}, this layer would incorporate Neural-Symbolic Systems, Few-Shot Learning capabilities, and Rule-Based Systems.

Neural-Symbolic Systems would combine the pattern-recognition capabilities of modern neural networks with explicit symbolic reasoning. This hybrid approach aims to bridge the gap between statistical language processing and cultural knowledge representation, implementing architectures that can handle both regular language patterns and cultural-specific expressions.

Few-Shot Learning mechanisms would be particularly crucial for preserving low-resource languages, implementing advanced meta-learning techniques that can effectively learn from limited examples. This component would address one of the key challenges in language preservation: the often limited availability of formal linguistic data.

Rule-Based Systems would complement these learning mechanisms by encoding explicit linguistic and cultural knowledge. These systems would implement community-validated rules and patterns, ensuring that technological solutions respect and preserve community-specific linguistic practices.

\subsubsection{Interface Layer}

The Interface Layer, as shown in Figure~\ref{fig:tech-ecosystem}, would provide the critical connection between communities and the technological system through three main components: Speech Processing, Text Processing, and Multimodal Interface systems. This layer would emphasize accessibility and cultural appropriateness in all interactions.

Speech Processing systems would be designed to handle the full spectrum of community language use, with particular attention to dialect variation and traditional speech patterns. We propose implementing adaptive acoustic models capable of learning from community input, ensuring that speech technologies respect and preserve local pronunciation patterns and speaking styles. These systems would be particularly crucial for languages with strong oral traditions \citep{Bird_2022}.

Text Processing capabilities would extend beyond standard character encoding to embrace the full range of community writing practices. The proposed system would support multiple scripts and writing systems, implementing flexible processing pipelines that can adapt to community-specific textual conventions. This component would be especially important for communities with unique writing traditions or those developing new writing systems.

Multimodal Interface systems would acknowledge that language expression often extends beyond speech and text. We propose implementing interfaces capable of handling various forms of linguistic and cultural expression, including gesture recognition for sign languages and multimodal communication patterns specific to particular communities. These interfaces would be designed to adapt dynamically to community preferences and usage patterns.

\subsubsection{Integration Layer}

The Integration Layer would provide the crucial infrastructure for ensuring system cohesion and community collaboration. As illustrated in Figure~\ref{fig:tech-ecosystem}, this layer would comprise Distributed Systems, API Framework, and Resource Protocols.

Distributed Systems would form the backbone of the technological infrastructure, implementing cloud-based processing capabilities that support collaborative development while maintaining community control. These systems would be designed to ensure that communities can maintain sovereignty over their linguistic data while benefiting from advanced processing capabilities \citep{Munk_etal_2022}.

The API Framework would establish standardized interfaces for component integration and interoperability. We propose implementing both RESTful and GraphQL interfaces, providing flexible integration options while maintaining strict data protection protocols. This framework would be designed to support both community-developed tools and integration with existing language technology systems.

Resource Protocols would implement sophisticated data sharing and protection mechanisms. These protocols would be designed to ensure that linguistic resources can be shared appropriately while respecting community rights and cultural protocols. The proposed system would include mechanisms for implementing community-defined access controls and cultural protection requirements.

\subsubsection{Preservation Layer}

The final layer of our proposed ecosystem, the Preservation Layer, would focus on ensuring the long-term sustainability and protection of linguistic and cultural resources. As shown in Figure~\ref{fig:tech-ecosystem}, this layer would include Archives, Version Control, and Access Management systems.

Archives would implement specialized digital preservation systems designed specifically for linguistic and cultural materials. We propose developing archival systems capable of maintaining the integrity of various data types – from traditional linguistic documentation to multimedia cultural records. These systems would implement advanced backup protocols while ensuring that archived materials remain accessible and usable by communities.

Version Control systems would track and manage changes in linguistic resources over time. These systems would be designed to document language evolution and variation, implementing sophisticated versioning protocols that can capture both formal linguistic changes and shifts in cultural usage patterns. This component would be particularly important for documenting living languages as they adapt and evolve.

Access Management would provide comprehensive security and access control mechanisms. We propose implementing flexible permission systems that can adapt to community protocols while ensuring appropriate scholarly and community access. These systems would be designed to protect sensitive cultural content while supporting language preservation and development efforts.

Through this layered ecosystem, we aim to create a technological framework that not only preserves linguistic data but actively supports community-driven language development and cultural preservation. Each layer has been designed to work in concert with the others, creating a cohesive system that maintains focus on meaning preservation and community empowerment throughout all levels of implementation.

\subsection{Integration and Validation Framework}

The proposed five-layer ecosystem requires a comprehensive framework for ensuring effective integration between layers and continuous validation of its alignment with community needs and meaning preservation goals. This section outlines our proposed approaches for integration mechanisms, validation processes, and sustainability strategies.

\subsubsection{Inter-Layer Integration Mechanisms}

The effectiveness of the proposed ecosystem would depend critically on seamless integration between its layers. We propose implementing what we term "vertical meaning preservation channels" – dedicated integration mechanisms that would ensure the consistent preservation of cultural meaning across all layers of the system.

Between the Knowledge Representation and Intelligence layers, we propose implementing bidirectional knowledge flow protocols. These would ensure that the symbolic representations of cultural knowledge effectively inform AI processing while allowing learning outcomes to enhance knowledge representations. This integration would be particularly crucial for maintaining what \citep{Marcus_Davis_2020} describes as the "semantic coherence" of the system.

The Intelligence and Interface layers would be connected through what we term "adaptive interaction bridges." These mechanisms would ensure that the system's intelligence capabilities directly inform interface adaptations while user interactions contribute to system learning. We propose implementing these bridges using flexible middleware that can evolve with community needs and technological capabilities.

The Integration and Preservation layers would work in concert through "heritage preservation protocols." These would ensure that all system integrations maintain strict adherence to community-defined preservation requirements while supporting system evolution. Following principles established by \citep{Crawford_2021}, these protocols would implement specific checks for cultural integrity at each integration point.

\subsubsection{Community Validation Processes}

We propose implementing a comprehensive validation framework that would operate at multiple levels of the ecosystem. This framework would extend beyond traditional technical validation to include what we term "cultural resonance verification" – systematic processes for ensuring that technological implementations align with community understanding and practices.

The proposed validation processes would include:

\begin{enumerate}
   \item Meaning Preservation Assessment: Systematic evaluation of how effectively the system preserves cultural meaning across all layers. This would implement specific protocols for tracking meaning integrity from knowledge representation through to user interfaces.
   
   \item Community Feedback Integration: Structured processes for gathering and incorporating community feedback at all system levels. We propose implementing what \citep{Sapignoli_2021} terms "recursive feedback loops" – mechanisms that ensure community input directly influences system evolution.
   
   \item Cultural Alignment Verification: Regular assessment of system behavior against community cultural practices and expectations. This would include specific protocols for identifying and addressing any divergence between system operation and cultural norms.
\end{enumerate}

\subsubsection{Meaning-Centered Evaluation Metrics}

Traditional technical metrics would be complemented by what we term "cultural performance indicators" – specialized metrics designed to assess how effectively the system preserves and transmits cultural meaning. These metrics would be developed in collaboration with communities and would include:

Cultural Authenticity Measures would assess how faithfully the system preserves and represents community knowledge and practices. These measures would implement specific indicators for evaluating the preservation of cultural context and meaning across system operations.

Community Empowerment Metrics would evaluate how effectively the system supports community agency in language preservation and development. Following frameworks proposed by \citep{Bird_2020}, these metrics would assess community control and participation across all system aspects.

Meaning Preservation Indices would track the system's effectiveness in maintaining semantic integrity across all layers. These indices would implement specific measures for evaluating how well cultural meaning is preserved through technological processing and representation.

\subsubsection{Sustainability and Evolution Strategies}

Long-term sustainability would be ensured through what we term "adaptive resilience frameworks" – systematic approaches to supporting system evolution while maintaining cultural integrity. These strategies would address both technical and cultural sustainability:

Technical Sustainability would be supported through modular architecture design and clear upgrade paths. 

Cultural Sustainability would be ensured through community ownership and control mechanisms. The proposed framework would implement specific protocols for maintaining community agency in system evolution, ensuring that technological advancement continues to serve cultural preservation goals.

Knowledge Transfer Mechanisms would support community capacity building for system maintenance and evolution. These mechanisms would implement structured approaches to sharing technical expertise while respecting and incorporating community knowledge systems.

Through these integrated frameworks and strategies, we aim to ensure that the proposed ecosystem remains both technically robust and culturally authentic over time, evolving in response to community needs while maintaining its core focus on meaning preservation and community empowerment.

\section{Challenges, Opportunities and Future Directions}

The proposed framework, while offering a comprehensive approach to language technology development, presents significant challenges in its implementation while simultaneously opening new opportunities for transforming how we approach linguistic diversity and cultural preservation. This section examines these dynamics and outlines future directions for research and development.

\subsection{Implementation Challenges}

The implementation of our proposed meaning-centered framework would face several interconnected challenges that need to be carefully addressed. At the knowledge representation level, we anticipate significant complexity in developing technical systems capable of capturing and preserving the nuanced ways in which communities construct and transmit meaning \citep{Bender_etal_2021}. The proposed integration of traditional knowledge systems with computational representations would require novel approaches that go beyond current technological paradigms.

Technical integration challenges would emerge primarily from the need to maintain semantic coherence across the five-layer ecosystem. The proposed hybrid systems, particularly in the Intelligence Layer, would need to balance advanced AI capabilities with community knowledge systems in ways that preserve cultural integrity \citep{Noble_2018}. This integration becomes particularly critical when dealing with low-resource languages and unique cultural contexts that may not align with conventional computational approaches.

Community engagement challenges would center on ensuring genuine participation while managing technical complexity. The proposed methodological framework would need to overcome potential barriers to community participation without compromising the sophistication of the technological solutions \citep{Hutchinson_etal_2021}. This includes addressing power imbalances between technology developers and speech communities, particularly in the context of technical decision-making and system evolution.

\subsection{Transformative Opportunities}

Despite these challenges, our framework opens significant opportunities for enhancing language preservation and community empowerment. By centering meaning and community agency, the proposed approach would enable the development of technologies that not only preserve linguistic forms but actively support the vitality of living languages. The five-layer ecosystem architecture would create opportunities for communities to maintain control over their linguistic and cultural heritage while leveraging advanced technological capabilities.

The proposed integration of traditional knowledge with modern technology would open new possibilities for preserving and sharing cultural wisdom. Our framework's emphasis on community-driven development would support the creation of sustainable models for language technology that serve community needs while respecting cultural integrity. Furthermore, the proposed methodological guidelines would facilitate knowledge transfer and capacity building, enabling communities to shape and evolve their technological solutions independently.

\subsection{Future Research and Development}

Looking forward, several critical paths emerge for advancing this paradigm shift in service of linguistic diversity and community empowerment. Future theoretical research should focus on deepening our understanding of meaning emergence within different cultural contexts, particularly in relation to technological integration. This would include developing new methodologies for studying the interaction between traditional knowledge systems and digital technologies.

The proposed technological ecosystem would need to evolve through the development of more flexible and culturally responsive approaches. We envision future development efforts focusing on creating adaptable tools and platforms that communities can meaningfully modify to serve their specific needs. This would include advancing the capabilities of all five layers while maintaining their interconnection and cultural alignment \citep{Miller_2022}.

Policy and governance frameworks would need to evolve to support this new paradigm. We propose developing comprehensive guidelines for ethical technology development in indigenous and minority language contexts, particularly focusing on protecting cultural intellectual property rights while promoting innovation. These frameworks should align with UNESCO's objectives for linguistic diversity and cultural preservation \citep{Bommasani_etal_2021}, while providing practical mechanisms for implementing the proposed meaning-centered approach to language technology development.

\section{Conclusions}

This paper presents a comprehensive framework for reimagining language technology development through the lens of AI Thinking and community agency. By challenging the current paradigm of technological homogenization, the proposed approach seeks to transform language technologies from extractive tools to collaborative solutions deeply rooted in cultural context.

The motivation emerges from a critical examination of foundation models' limitations, which simultaneously offer unprecedented computational capabilities and pose fundamental risks to linguistic diversity. The proposed meaning-centered approach repositions community agency as the core driver of technological innovation, recognizing the complex interplay between technological advancement and cultural preservation.

The AI Thinking framework provides a theoretical foundation for this transformation, establishing meaning as the central axis of technological development. Through a hierarchical structure integrating core principles, structural dimensions, and operational skills, the framework offers a systematic approach to bridging cultural understanding with technological innovation.

The proposed five-layer technological ecosystem translates these theoretical principles into a practical implementation strategy. Each layer is designed to maintain community agency while leveraging advanced technological capabilities, presenting a nuanced path that neither rejects technological progress nor compromises cultural integrity.

Significant challenges remain. Technical complexities include developing systems capable of capturing nuanced cultural meanings, creating hybrid approaches that synthesize AI capabilities with community knowledge, and ensuring semantic coherence across technological ecosystems. Equally critical are procedural challenges related to community participation, cultural authenticity, and genuine technological co-creation.

Governance and geopolitical considerations demand sophisticated frameworks addressing intellectual property, data sovereignty, power dynamics, and sustainable development models. These challenges, however, simultaneously present opportunities for more equitable, culturally responsive technological innovation.

The framework's future success depends on sustained commitment from diverse stakeholders: technology developers willing to embrace transformative paradigms, empowered communities actively shaping technological solutions, and adaptive governance supporting meaningful collaboration. Only through such collective endeavor can we envision technological advancement that genuinely serves and enriches cultural diversity.

\bibliography{referencias}

\end{document}